\documentclass[letterpaper]{article}
\usepackage{aaai}
\usepackage{times}
\usepackage{helvet}
\usepackage{courier}
\usepackage{graphicx}
\usepackage{comment}
\usepackage{multirow}
\usepackage{amsmath}
\usepackage{caption}
\usepackage{subcaption}
\usepackage{hyperref}
\usepackage{xcolor}
\begin{document}
 
\title{Towards Generating Large Synthetic Phytoplankton Datasets \\for Efficient Monitoring of Harmful Algal Blooms}
\author{Nitpreet Bamra\textsuperscript{1},
Vikram Voleti\textsuperscript{2,3},
Alexander Wong\textsuperscript{1,2},
Jason Deglint\textsuperscript{1,2}\\
\textsuperscript{1}{University of Waterloo},\\ 
\textsuperscript{2}{Blue Lion Labs},\\
\textsuperscript{3}{Mila, University of Montreal}\\
\{nbamra, jdeglint\}@uwaterloo.ca}

\maketitle
\begin{abstract}
\begin{quote}
Climate change is increasing the frequency and severity of harmful algal blooms (HABs), which cause significant fish deaths in aquaculture farms. This contributes to ocean pollution and greenhouse gas (GHG) emissions since dead fish are either dumped into the ocean or taken to landfills, which in turn negatively impacts the climate. Currently, the standard method to enumerate harmful algae and other phytoplankton is to manually observe and count them under a microscope. This is a time-consuming, tedious and error-prone process, resulting in compromised management decisions by farmers. Hence, automating this process for quick and accurate HAB monitoring is extremely helpful. However, this requires large and diverse datasets of phytoplankton images, and such datasets are hard to produce quickly. In this work, we explore the feasibility of generating novel high-resolution photorealistic synthetic phytoplankton images, containing multiple species in the same image, given a small dataset of real images. To this end, we employ Generative Adversarial Networks (GANs) to generate synthetic images. We evaluate three different GAN architectures: ProjectedGAN, FastGAN, and StyleGANv2 using standard image quality metrics. We empirically show the generation of high-fidelity synthetic phytoplankton images using a training dataset of only 961 real images. Thus, this work demonstrates the ability of GANs to create large synthetic datasets of phytoplankton from small training datasets, accomplishing a key step towards sustainable systematic monitoring of harmful algal blooms.
\end{quote}
\end{abstract}
\frenchspacing
\section{Introduction}
\label{sec:intro}

When different phytoplankton and algae species grow uncontrollably, they can form harmful algal blooms (HABs). These HABs can produce lethal toxins and hypoxic ``dead" zones, causing catastrophic impacts on various industries such as aquaculture and real estate~\cite{HABs2}, as well as negatively impacting wildlife and the environment~\cite{HABs}. Further, research shows that climate change is increasing the frequency and severity of these HABs \cite{WELLS2020101632}. For example, in 2016, a HAB outbreak in Chile killed over 27 million farmed trout and salmon~\cite{MONTES201855}. Furthermore, from 2015-2019, farmed salmon deaths due to fatal diseases caused by HABs increased by 27.8\% from 41.3 to 52.8 million~\cite{salmon}. When millions of fish die due to HABs and other diseases, they are disposed of by being dumped back into the ocean, causing ocean pollution, or being brought to landfills, causing increased greenhouse gas (GHG) emissions~\cite{ARMIJO2020110603,Bustos2021}. Furthermore, as GHG emissions increase, the surface temperature and acidity of water bodies increase simultaneously, creating an ideal environment for HABs to grow, thereby leading to an endless cycle of increasing GHG emissions~\cite{impactsofCC}.

A promising solution to mitigate this problem is quicker and more consistent monitoring of HABs. This would improve farm yields, as well as reduce ocean pollution and GHG emissions. However, rapid identification is challenging due to the extremely time-consuming, tedious and error-prone process for a taxonomist to manually analyze and classify these algae species~\cite{HABs3}.

Given digital microscopy and novel deep learning methods, it is possible to improve both the speed and accuracy of current algae detection methods~\cite{app10176033}. However, such methods require the collection of large and diverse phytoplankton datasets, which is both expensive and tedious, thereby limiting the effectiveness of current HAB monitoring as shown in the ``Current Method'' section of Figure \ref{fig:overview}. Moreover, it is important that such datasets match real images as far as possible, so they should ideally include multiple phytoplankton species in the same image.

In this work, we propose a framework to generate high-fidelity and diverse synthetic datasets of phytoplankton, motivated by the potential to expedite, improve, and standardize algae and phytoplankton detection. This is shown in the ``Proposed Method'' section of Figure \ref{fig:overview}. We utilize Generative Adversarial Networks (GANs) to generate novel synthetic images, but in principle any image generation technique could be used. Our main contributions in this work are: (1) we explore three state-of-the-art (SOTA) GAN architectures for generating synthetic datasets of image samples containing multiple phytoplankton species, (2) we evaluate the generated images using standard image quality metrics, and (3) we validate the novelty of generated images by checking for memorization of the training images.

\begin{figure}[t]
    \centering
    \scalebox{0.85}{
        \includegraphics[width=\columnwidth]{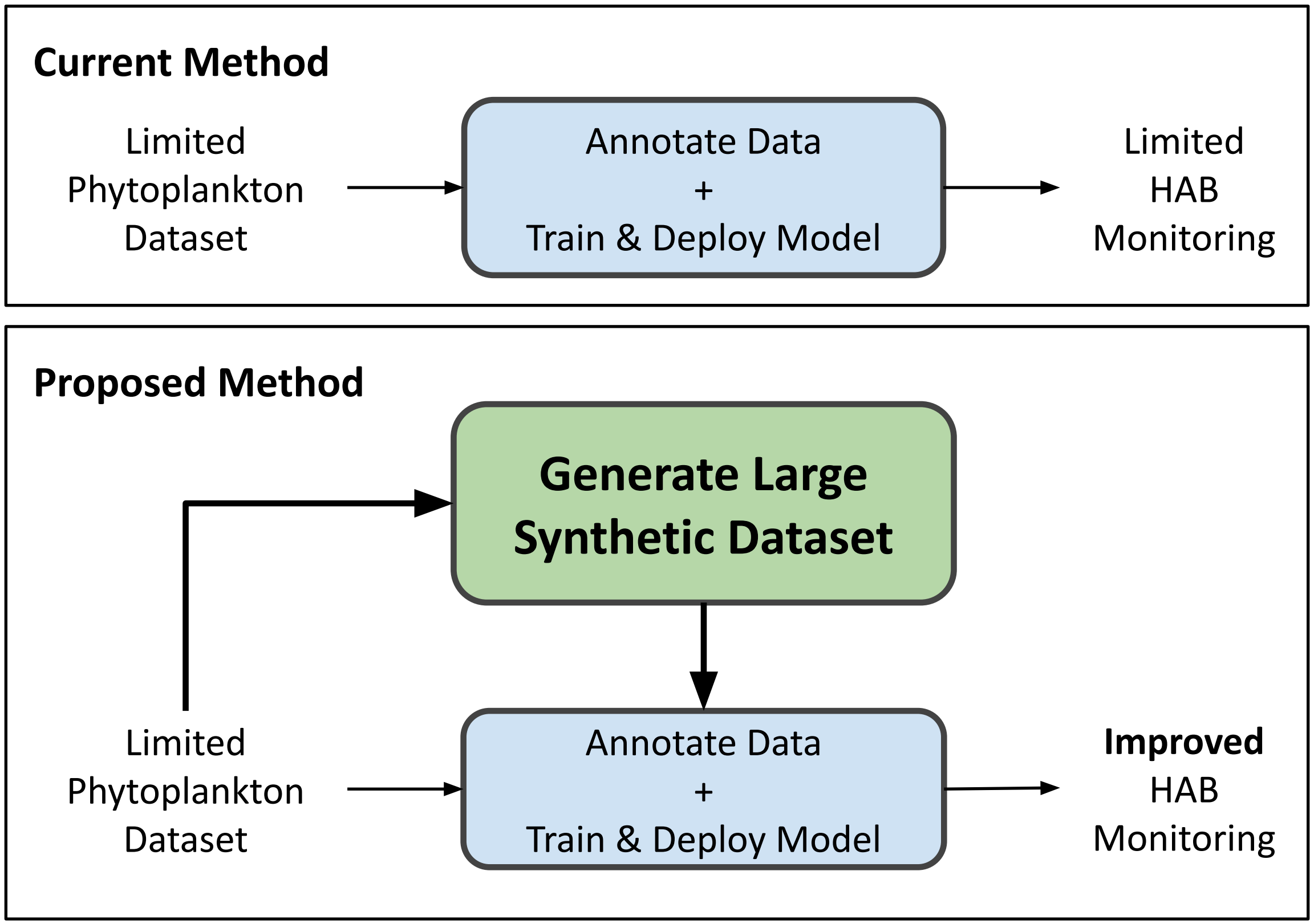}
        }
        \caption{The current (top) and proposed (bottom) deep learning algae detection methods. Our proposed framework leads to improved HAB monitoring, the initial step being the generation of a large and synthetic algae dataset.}\
        \label{fig:overview}
\vspace{-4mm}
\end{figure}

\begin{figure*}[!th]
        \begin{subfigure}[b]{0.25\textwidth}
                \centering
                \includegraphics[width=.943\linewidth]{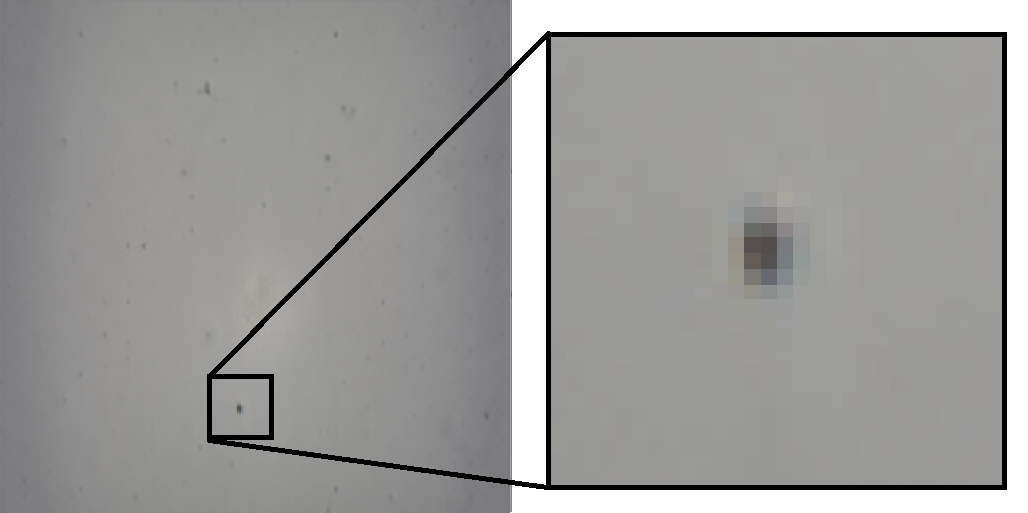}
                \includegraphics[width=.943\linewidth]{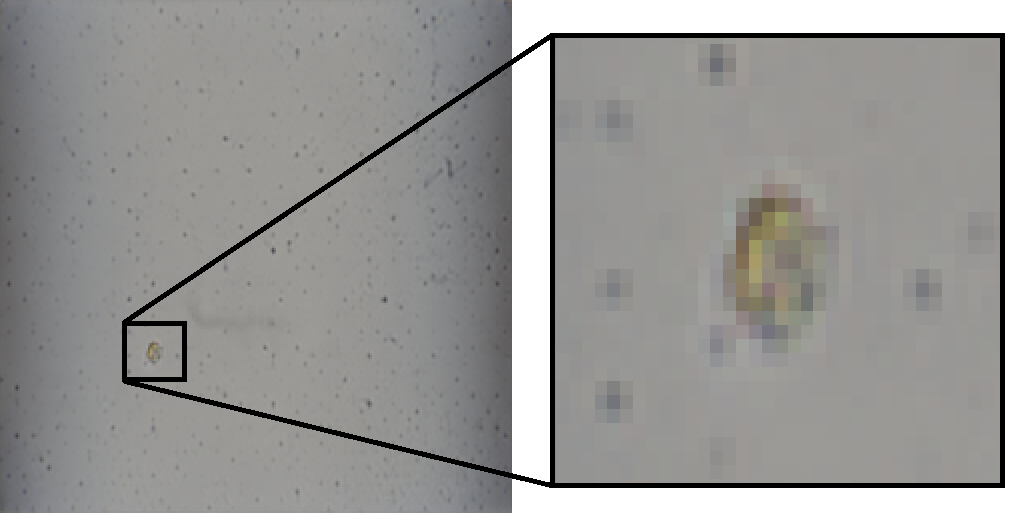}
                \caption{ProjectedGAN Images}
                \label{fig:projGAN}
        \end{subfigure}%
        \begin{subfigure}[b]{0.25\textwidth}
                \centering
                \includegraphics[width=.943\linewidth]{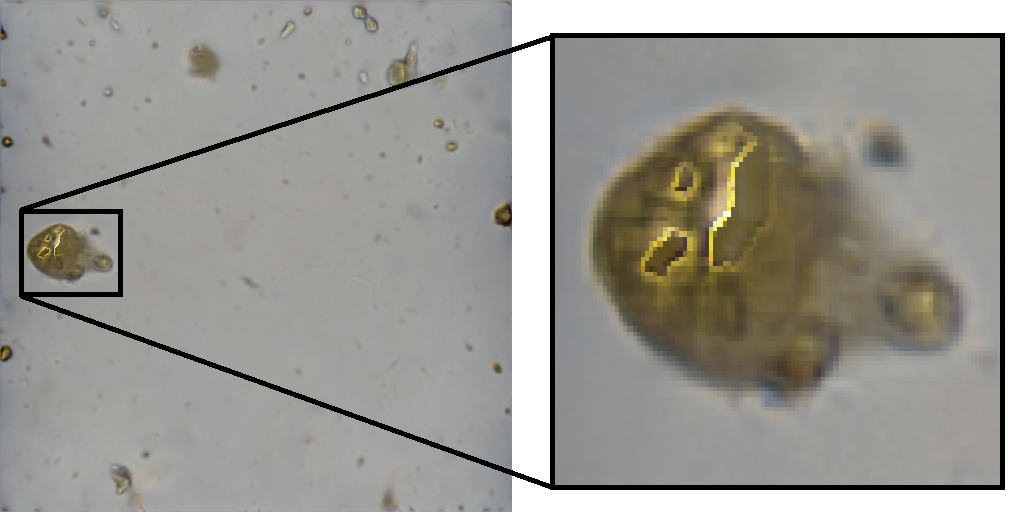}
                \includegraphics[width=.943\linewidth]{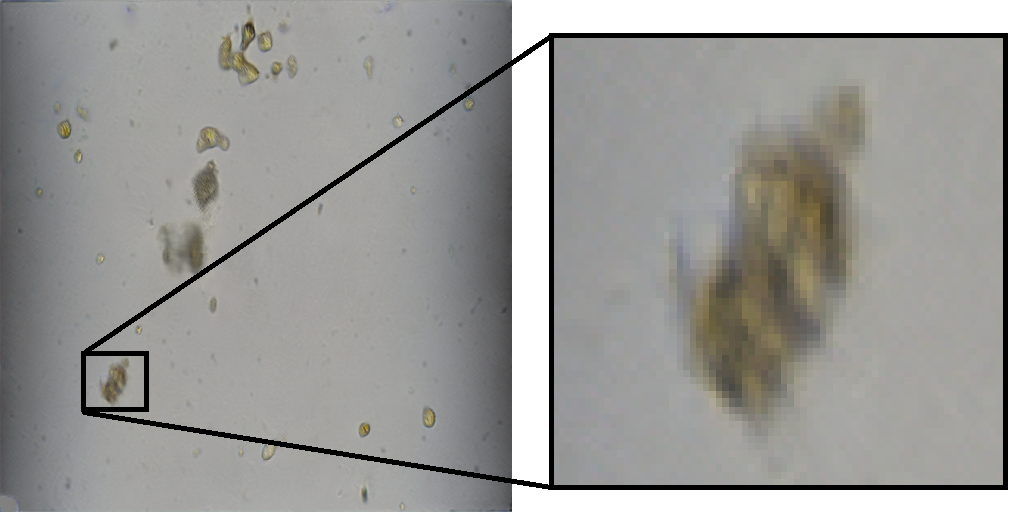}
                \caption{FastGAN Images}
                \label{fig:fastGAN}
        \end{subfigure}
        \begin{subfigure}[b]{0.25\textwidth}
                \centering
                \includegraphics[width=0.943\linewidth]{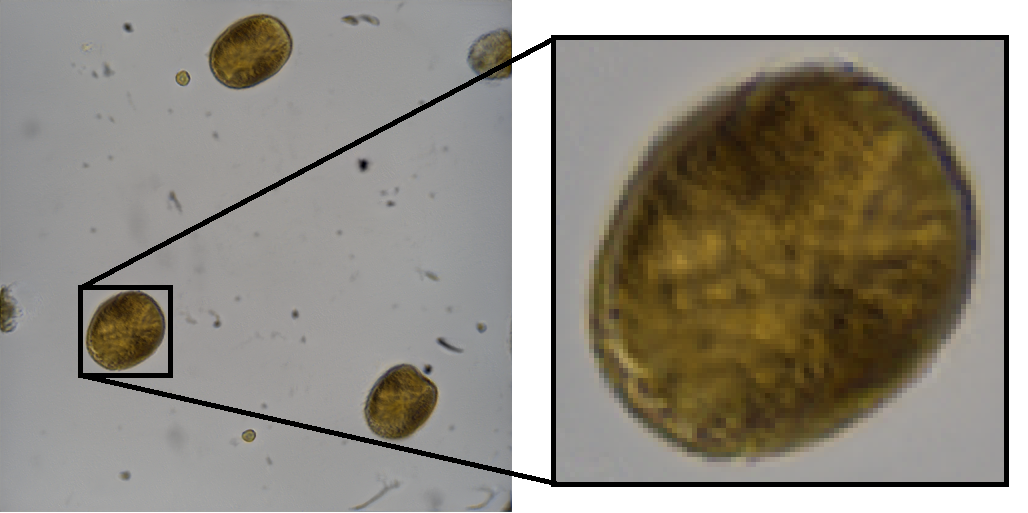}
                \includegraphics[width=0.943\linewidth]{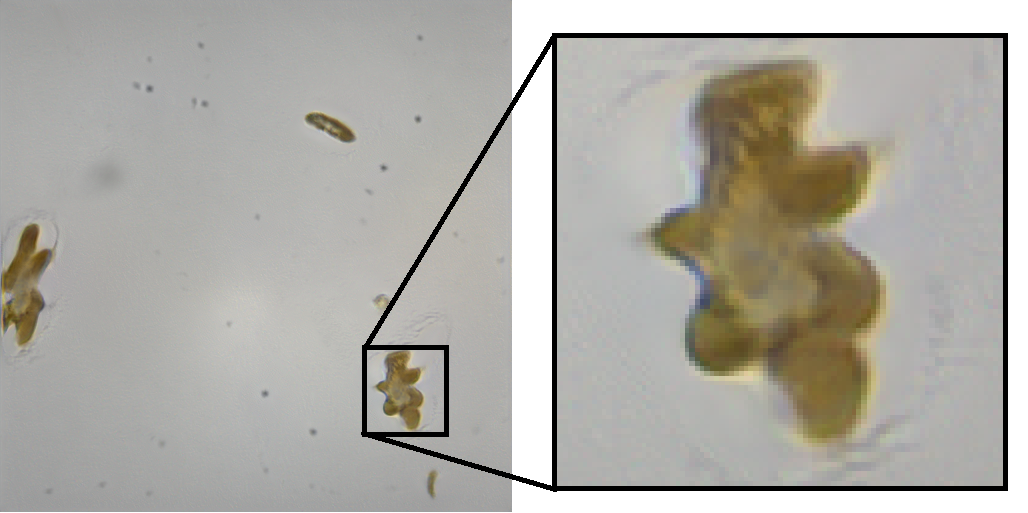}
                \caption{StyleGANv2 Images}
                \label{fig:stylegan}
        \end{subfigure}%
        \begin{subfigure}[b]{0.25\textwidth}
                \centering
                \includegraphics[width=.91\linewidth]{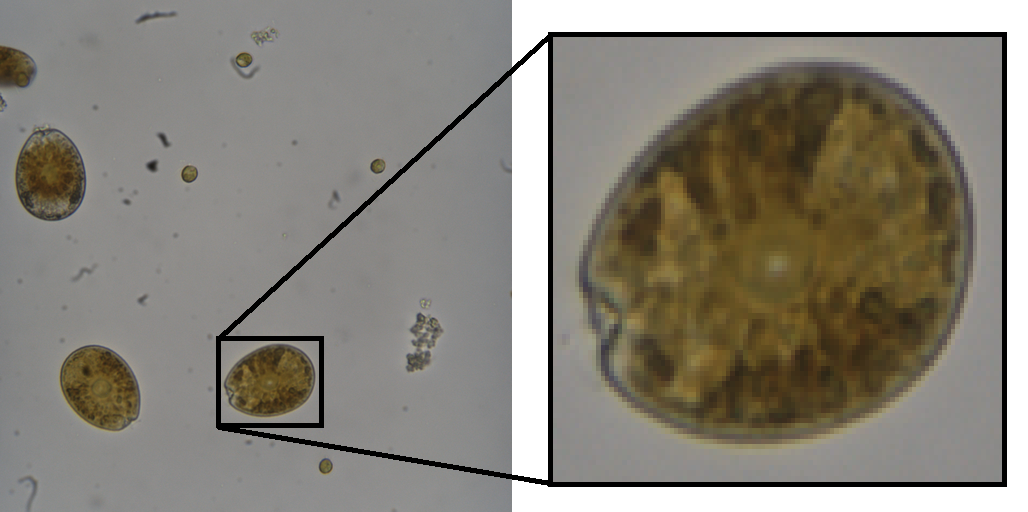}
                \includegraphics[width=.91\linewidth]{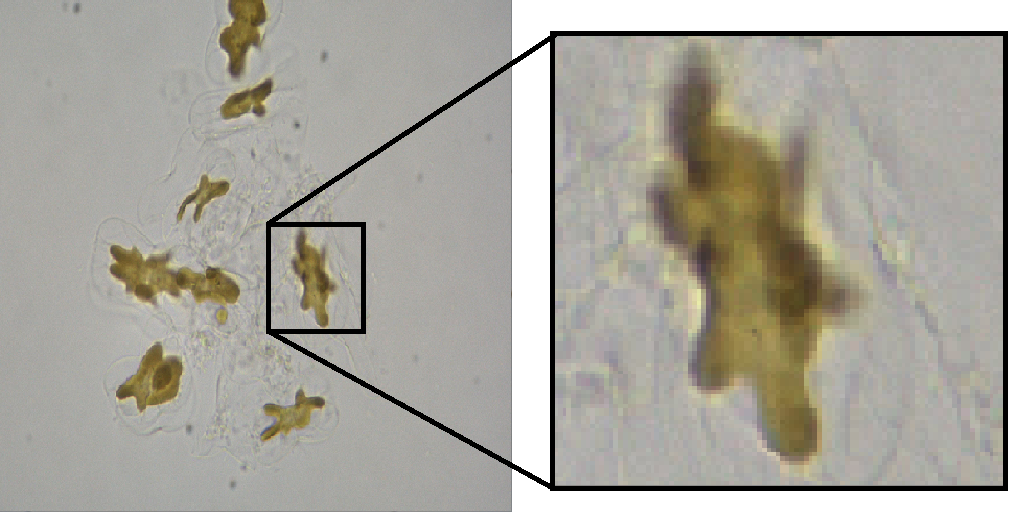}
                \caption{Real Images}
                \label{fig:real}
        \end{subfigure}%
        \caption{(a)-(c) Generated images from the three different GANs, as well as (d) real images from the training dataset. Observe that when comparing to real images (d), ProjectedGAN has the least realistic generated images (a), the FastGAN has slightly more realistic generated images (b), and the SytleGANv2 has the most realistic images (c). For each whole 1024x1024 image (left), an enhanced image of a single algae specimen is provided (right).}\label{fig:Images}
\vspace{-5mm}
\end{figure*}
\section{Related Works}

GANs are some of the most effective machine learning frameworks for generating synthetic outputs from random noise inputs~\cite{goodfellow2014generative}. A GAN model contains two sub-models: a generator $G$ and a discriminator $D$. The generator works to create realistic synthetic images from a random noise input, and the discriminator deciphers whether an image is real or generated. To improve the quality of the generated images, the two sub-models compete against each other in a two-player min-max game, the objective is calculated using Equation \ref{eq:GANeq}: 
\begin{multline}
\label{eq:GANeq}
\mathop{min}_{{G}} \mathop{max}_{{D}} (E_x[\log D(x)]+ E_z[log(1-D(G(z)))])
\end{multline}

In Equation \ref{eq:GANeq}, $D(x)$ and $D(G(z))$ denote the probability that an instance of real data $x$ or generated images $G(z)$ from noise instance $z$ is from the real dataset or the generator. $E_x$ and $E_z$ denote the expected values from all real data inputs to the discriminator and random noise inputs to the generator, respectively. 

A study published in 2017~\cite{8296402} suggested that convolutional neural networks tend to be biased towards larger specimen classes. It then proposed to use the ConditionalGAN model~\cite{mirza2014conditional} to generate synthetic images of plankton species to address the hardships of species generation given few per-class examples.

Another study published in 2021 ~\cite{jmse9060636} explores the use of the CycleGAN model~\cite{zhu2017unpaired} to address the imbalance within the classes of plankton species by generating more of the low-class species. The CycleGAN architecture allows for image-to-image translation by learning and interchanging the mapping features of the two images.

Lastly, a 2022 study from Inje University~\cite{9798730} focused on
% algae detection and classification methods by 
using the DCGAN model ~\cite{DCGAN} to generate microscopic images of phytoplankton to enhance the size of their dataset. The DCGAN model is an extension of the original GAN model, the main difference being the use of deep convolutional neural networks for both the discriminator and generator, instead of fully connected layers. This study used a dataset of 400 single-organism microscopic images of 4 different algae species. They fed these 400 images to the DCGAN, using it as an ``advanced augmentation'' tool to successfully double their dataset.

In contrast, our investigation focuses on generating a novel high-quality synthetic dataset of 11 organism types, in which each image contains multiple algae species. Furthermore, we explore three different SOTA GAN architectures, and generate high-resolution images.

\section{Methodology}

\subsection{GAN Architectures}
Three GAN models that have proven to produce high-fidelity synthetic images consistently are ProjectedGAN~\cite{ProjectedGAN}, FastGAN~\cite{FastGAN}, and StyleGANv2~\cite{stylegan2}. 

\noindent
\\
\textbf{Projected GAN. }
The ProjectedGAN~\cite{ProjectedGAN} works by projecting features from both generated and real images into a pre-trained feature space instead of the standard input space. The min-max equation of ProjectedGAN differs from that of vanilla GAN by introducing projections of different feature vectors $P_{\iota}(x)$ into the discriminator's original input space, and is calculated using Equation \ref{eq:ProjGAN}:
\begin{multline}
\label{eq:ProjGAN}
\mathop{min}_{{G}} \mathop{max}_{{D_\iota}} \Sigma_{\iota\epsilon\zeta} (E_x[\log D_\iota (P_\iota(x))] \\
+  E_z[log(1-D_\iota(P_\iota(G(z))))])
\end{multline}
In Equation \ref{eq:ProjGAN}, $D_{\iota}$ denotes a set of independent discriminators which operate on distinct projected features and $\Sigma_{\iota\epsilon\zeta}$ denotes the summation of all components in the equation. 

\noindent
\\
\textbf{FastGAN. }
The FastGAN~\cite{FastGAN} model reduces the significant computational needs and the large number of training images required for a GAN to train, by incorporating a Skip-Layer channel-wise Excitation (SLE) module. The SLE module is calculated using Equation \ref{FastGANEq}, where $y$ is the output feature map of the SLE model, $W_i$ is the weight which needs to be learned by the model, and $x_{low}$/$x_{high}$ represent the low/high resolution feature maps, respectively.
\begin{equation}
\label{FastGANEq}
y = F(x_{low},(W_i)) * x_{high}
\vspace{-3mm}
\end{equation}   

\noindent
\\
\textbf{StyleGANv2. }
StyleGANv2~\cite{stylegan2} is one of the SOTA GAN models that consistently produces high-fidelity synthetic images without using pre-trained feature spaces. This is achieved by first transforming the input noise vector into an `` intermediate latent code" using a mapping network. These intermediate codes represent multiple distinct ``styles", which allows for the GAN to generate images with multiple layers of details, ranging from course details like algae shape and orientation, to fine details like algae flagella and cilia. These ``styles" are then processed through an adaptive instance normalization (AdaIN) process, as explained in Equation \ref{styleganEQ}:
\begin{equation}
\label{styleganEQ}
AdaIN(x_i,y)=y_{s,i} \frac{x_i - \mu(x_i)}{\sigma(x_i)} + y_{b,i} 
\end{equation}
In Equation \ref{styleganEQ}, $x_i$ represents a feature map, $y_b$ and $y_s$ represent corresponding scalar component from the respective style and $\mu$/$\sigma$ represent scalable factors for input normalization ~\cite{AdaIN}. 

\begin{table*}[h!]
\caption{Image quality and computational metrics from each GAN model. The SytleGANv2 consistently yields the lowest FID and KID scores (see bold values) supporting that the SytleGANv2 generated the best synthetic images.}
\centering
\resizebox{2.12\columnwidth}{!}{
\renewcommand{\arraystretch}{1}
\begin{tabular}{ |c|c|c|c|c|c|c|c|c|c|c|c|} 
\hline
\textbf{GAN Model} & \textbf{Data Size} & \textbf{Resolution} & \textbf{Iterations} & \textbf{Batch Size} & \textbf{Training Time} & \textbf{FID} & \textbf{KID} & \textbf{GPU model} \\ [0.7ex] 
\hline
\multirow{2}{6em}{ProjectedGAN} & \ 1922 & 256x256 & 1,427,400 & 64 & 04d 00h 45m  & \textbf{113.107} & \textbf{0.050} & GTX TITAN X \\
& \ 1922 & 256x256 & 1,008,000 & 64 & 02d 20h 33m  & 226.317 & 0.093  &  GTX TITAN X\\
\hline
\multirow{4}{3.5em}{FastGAN} & \ \ 961 & 1024x1024 &\ \ \ \ \ 50,000 & \ 8 & \ \ \ \ \ \ 21h 33m & 186.690 & 0.163&  GTX TITAN X \\ 
  & \ 9610 & 1024x1024 & \ \ \ \ \ 25,000 & \ 8 &\ \ \ \ \ \ 10h 46m &\textbf{130.201}&0.090& GTX TITAN X\\ 
 &  \ 9610 & 1024x1024 & \ \ \ \ \ 50,000 & \ 8 & \ \ \ \ \ \ 16h 00m & 166.062 & \textbf{0.083}& GTX TITAN X\\ 
 & \ 9610 & 3208x2200 & \ \ \ \ \ 50,000 & \ 8 & \ \ \ \ \ \ 08h 02m & 190.040 & 0.115& $2\mathbf{x}$ RTX 2080 Ti\\
\hline
\multirow{2}{5em}{StyleGANv2} & \ 1922 & 512x512  & 3,000,000 & \ 4  & 02d 14h 58m & \textbf{29.330} & \textbf{0.014} &$2\mathbf{x}$ RTX 2080 Ti \\
 & 19220 & 1024x1024  & \ \ \ 824,000 & \ 4  & 08d 04h 48m  & 43.423 & 0.015 & GTX TITAN X \\
\hline
\end{tabular}
}
\label{Table:metric}
\centering
\end{table*}

\subsection{Image Quality Metrics}
Overall, since each GAN model has distinct advantages and disadvantages, the effectiveness of each GAN can be evaluated using both image quality and computational metrics.

In terms of computational statistics, the metrics measured in this investigation are: dataset size, dataset image resolution, number of iterations, batch size and time to train. In terms of image quality, the two metrics evaluated in this investigation are Fréchet inception distance (FID) and Kernel-inception distance (KID).

In this investigation, some industry standard image quality metrics, such as Structure similarity (SSIM)~\cite{SSIM} and Mean-squared error (MSE)~\cite{mse}, are not appropriate since these metrics directly compare the corresponding pixel values of a real image and generated image. This is because the generated plankton images should have plankton in various locations across the image. FID and KID compare the feature vectors between the generated and real images, providing better insight into the similarities between the two image distributions. Lower FID and KID numbers indicate more similarity in the image distributions.

\noindent
\\
\textbf{Fréchet Inception Distance.}
The FID metric~\cite{fid} compares the distribution of generated images with the distribution of authentic images by calculating the difference between the feature vectors for the generated and real images as shown in Equation \ref{FIDeq}: 
\begin{multline}
\label{FIDeq}
d^2((m,C),(m_w,C_w)) = {||m-m_w||}_2^2 \\ 
+ Tr(C+C_w-2(CC_w)^{1/2}
\end{multline}
In Equation \ref{FIDeq}, $m$ and $C$ are the mean and covariance feature polynomial from the generated images, $m_w$ and  $C_w$ is the mean and covariance feature polynomial acquired from the training dataset, and $Tr$ is the linear trace function.

\noindent
\\
\textbf{Kernel Inception Distance.}
The KID metric~\cite{kid} is similar to FID, with the main difference being that KID measures the squared maximum mean discrepancy $MMD$ between the real image feature representations $f_{real}$ and fake image feature representations $f_{fake}$, as shown in Equation \ref{kid}: 
\begin{equation}
\label{kid}
KID = MMD(f_{real},f_{fake})^2
\end{equation}
\subsection{Datasets}
The dataset used in this investigation contains eleven different phytoplankton species, including \textit{Entomoneis paludosa},   \textit{Alexandrium catenella}, \textit{Porphyridium purpurem}, \textit{Navicula sp.}, \textit{Heterosigma akashiwo}, \textit{ Alexandrium ostenfeldii}, \textit{ Porphyridium purpureum}, \textit{Dolichospermum}, \textit{ Phaeodactylum tricornutum M1} and \textit{Phaeodactylum tricornutum M2}. 
The dataset had 961 distinct microscope specimen photos, each at a resolution of 3208x2200 pixels. These images were then further manipulated to create two different datasets, one consisting of all 961 images being center-cropped and the other consisting of the 961 images being randomly cropped ten times, to create a larger dataset of 9610 images. Furthermore, when testing the ProjectedGAN and StyleGANv2 models, each image within the two datasets was flipped on the x-axis, thereby doubling the dataset to 1922 and 19220 images, respectively. Lastly, the GANs in this investigation were trained on two different PCs, one with a single 12GB Nvidia GeForce GTX TITAN X GPU and the other with two 11GB GeForce RTX 2080 Ti GPUs.
\section{Results \& Discussion}
\textbf{Qualitative Comparison.} Figure \ref{fig:Images} shows a visual comparison of the real algae images and the synthetic generated algae for each GAN model trained in this investigation. Based on visual inspection, conclusions can be made about each model's effectiveness. Figure \ref{fig:projGAN} shows the results from the ProjectedGAN model. All the generated images are entirely blank, implying that the GAN had experienced mode collapse. In contrast, the generated images from FastGAN in Figure \ref{fig:fastGAN} are somewhat similar to the real images in Figure \ref{fig:real}, however, the noise within the generated images is still apparent. Lastly, Figure \ref{fig:stylegan} shows the results from the StyleGANv2 model, and it is clear that the generated images are very similar to the real images.

\noindent
\\
\textbf{Image Quality Metric Comparison.} Beyond a strictly visual comparison, both the image quality and computational metrics can be seen in Table \ref{Table:metric}. When comparing the image quality metrics, the StyleGANv2 model yielded the lowest FID and KID scores by a significant margin compared to the other GAN models. The lowest FID score from the StyleGANv2 was 29.330, compared to 130.201 and 113.107 from the FastGAN and ProjectedGAN, respectively. The lowest KID score from the  StyleGANv2 was 0.014 compared to 0.083 from the FastGAN and 0.050 from the ProjectedGAN. Therefore, it can be concluded that the quality of generated images from the StyleGAN was about four to five times better than that of the other tested GANs.

\noindent
\\
\textbf{Computational Metric Comparison.} Although the StyleGANv2 model can generate the highest fidelity images; it is at the expense of longer training times and more extensive computational requirements. For example, it took over two days to train the StyleGAN2 model for three million iterations using a 512x512 resolution dataset on a dual RTX 2080 Ti and over eight days to train for 824,000 iterations using a 1024x1024 resolution dataset on a single GTX TITAN X GPU. Conversely, the FastGAN model trains the fastest but converges at the highest FID and KID values. Lastly, the ProjectedGAN model  took around two to four days to train for about 1,000,000 iterations on the GTX GPU; however, in both instances, the GAN had clearly failed during training.
\subsection{Investigating Potential GAN Failure}
\begin{figure}[]
        \begin{subfigure}[b]{0.24\textwidth}
                \centering
                \includegraphics[width=.45\linewidth]{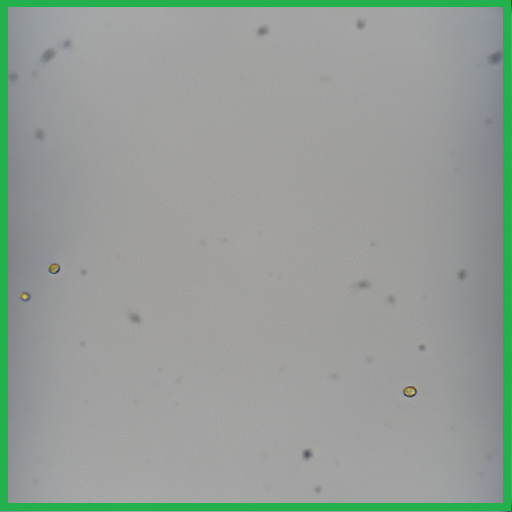}
                \includegraphics[width=.45\linewidth]{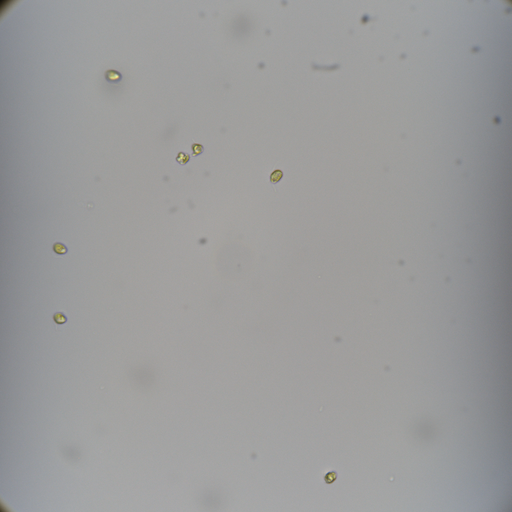}
                \includegraphics[width=.45\linewidth]{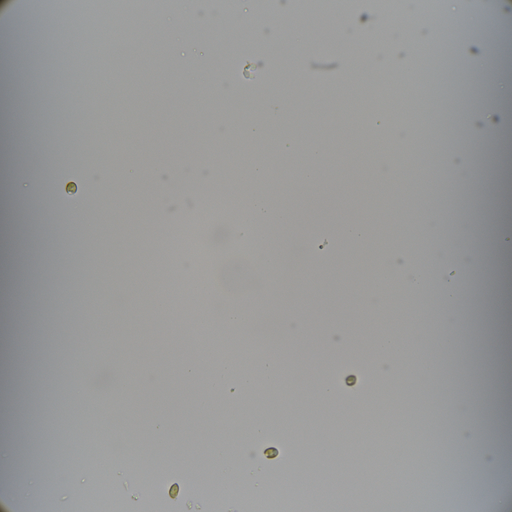}
                \includegraphics[width=.45\linewidth]{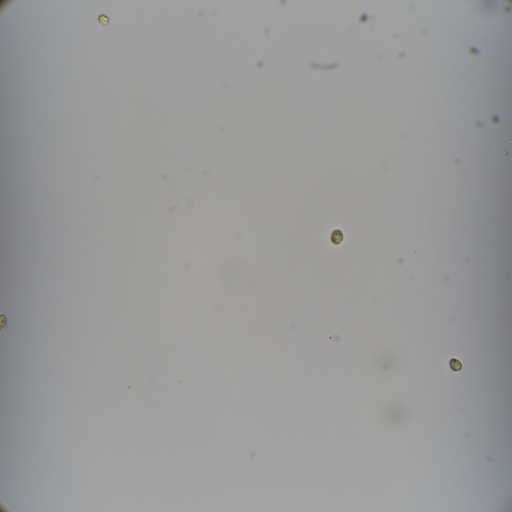}
                \caption{Pixel space}
                \label{fig:Pixelspace}
        \end{subfigure}%
        \begin{subfigure}[b]{0.24\textwidth}
                \centering
                \includegraphics[width=.45\linewidth]{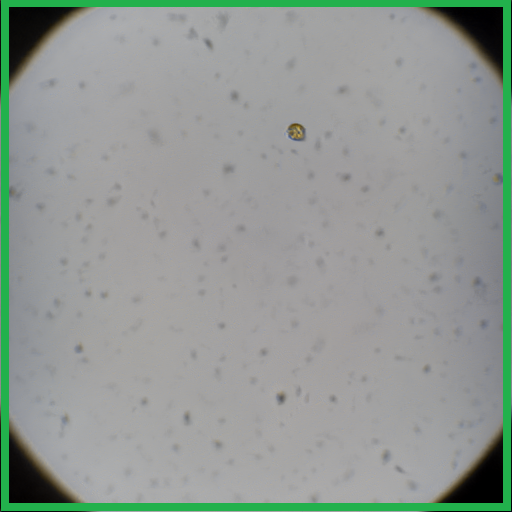}
                \includegraphics[width=.45\linewidth]{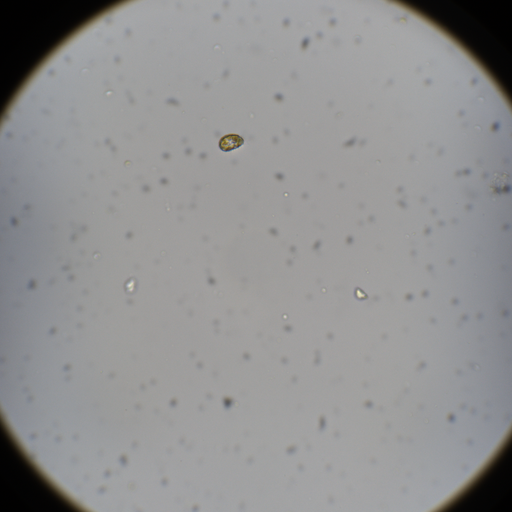}
                \includegraphics[width=.45\linewidth]{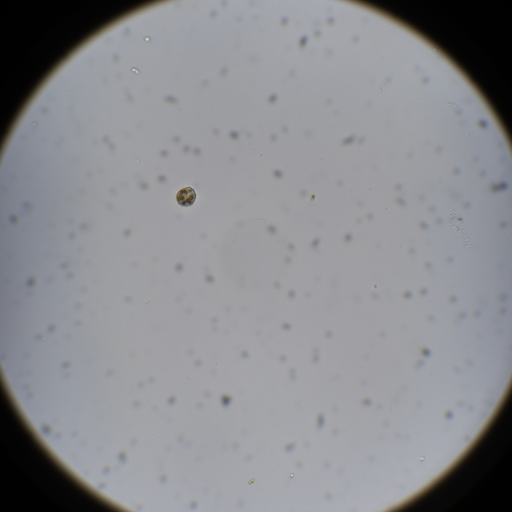}
                \includegraphics[width=.45\linewidth]{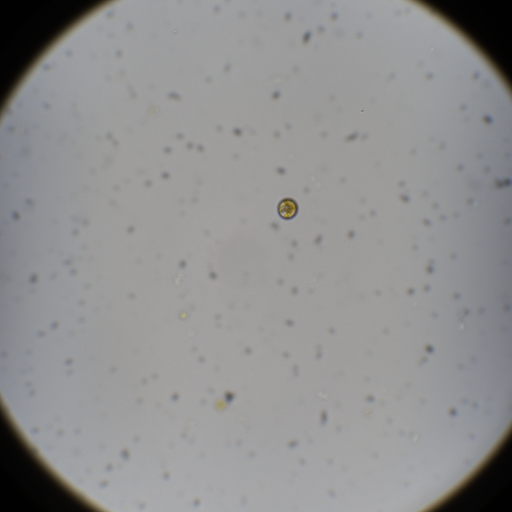}
                \caption{Feature space}
                \label{fig:Featurespace}
        \end{subfigure}
        \caption{The generated image from StyleGANv2 can be seen on the top left, and the respective three nearest neighbors can be seen in both (a) the pixel and (b) the feature space. This illustrates that the SytleGANv2 is not memorizing the dataset but generating novel images.}\label{fig:NN}
\vspace{-5mm}
\end{figure}
Since the StyleGANv2 yielded the best results in this investigation, the model was further tested to ensure that images generated by the GAN are novel. When working with GAN models trained with a limited number of training images, the model may potentially memorize images from the dataset instead of generating novel images. Therefore, we checked the nearest neighbours of generated images in the real dataset, nearness being measured in (a) pixel space and (b) feature space. We provide one example each in Figure \ref{fig:NN}, the nearest-neighbour analysis of the StyleGANv2 model shows that the synthetic images are indeed novel.

\section{Conclusions \& Future Works}
\vspace{-1mm}

This work provides a framework for integrating synthetic datasets into the critical application of HAB monitoring. We investigate the effectiveness of GANs in generating high-resolution novel photorealistc synthetic phytoplankton images from a small dataset of real images. We explored three SOTA GAN architectures, and found StyleGANv2 to be quite reliable qualitatively and quantitatively.

Given the outcomes of this paper, the immediate future work includes (1) labelling the real and generated datasets, and then (2) training a classifier with the original and the combined original and new datasets to evaluate model performance.
Finally, careful consideration must be taken on how these models are deployed in the field to ensure reliable and consistent predictions are provided to end users, such as aquaculture farmers. This will help mitigate the impacts of climate change by contributing a key step towards quick and accurate HAB monitoring.

\section{Acknowledgements}
This research was funded by the Waterloo AI Institute and the Mitacs Accelerate Program. The dataset was provided by Blue Lion Labs, and the computing resources were provided by the Vision and Image Processing (VIP) Lab at the University of Waterloo.

\frenchspacing
\bibliography{ref.bib}
\bibliographystyle{aaai}
\pagebreak
\section{Appendix}
\subsection{Appendix A : More Generated Images}
\begin{figure}[h]
    \centering
    \scalebox{0.9}{
        \includegraphics[width=\columnwidth]{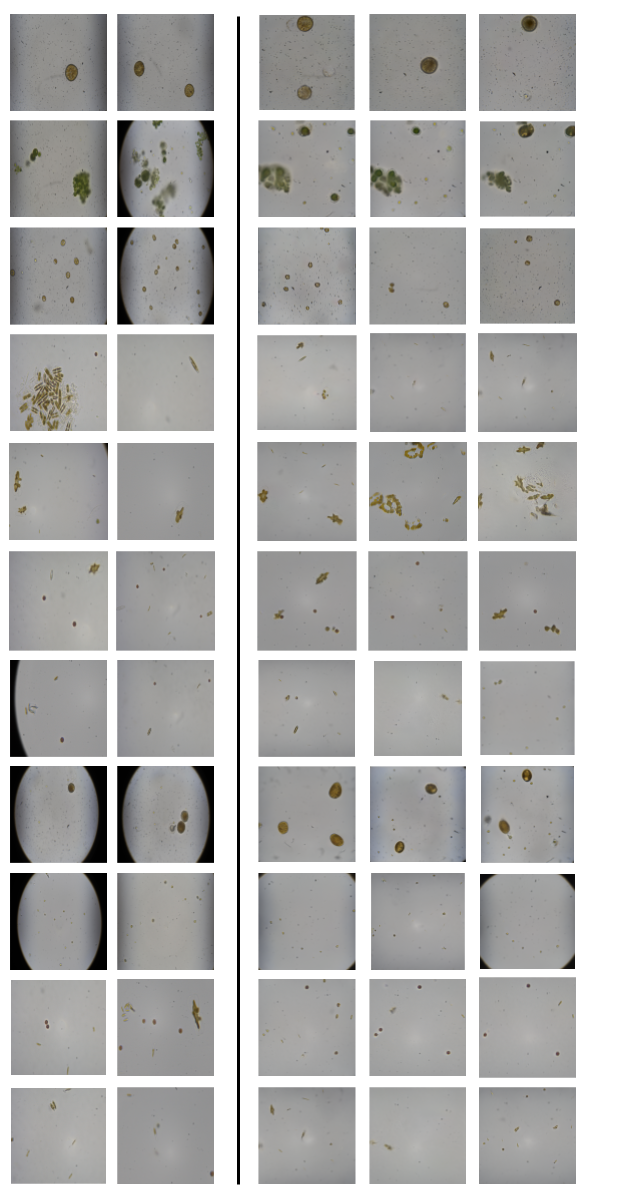}
        }
        \caption{The first two columns show real images, the last three columns show images generated by our trained StyleGANv2 model.}\
        \label{fig:overview}
\end{figure}

\pagebreak

\subsection{Appendix B : More Nearest Neighbor Comparisons}
\begin{figure}[h]
    \centering
        \includegraphics[width=\columnwidth]{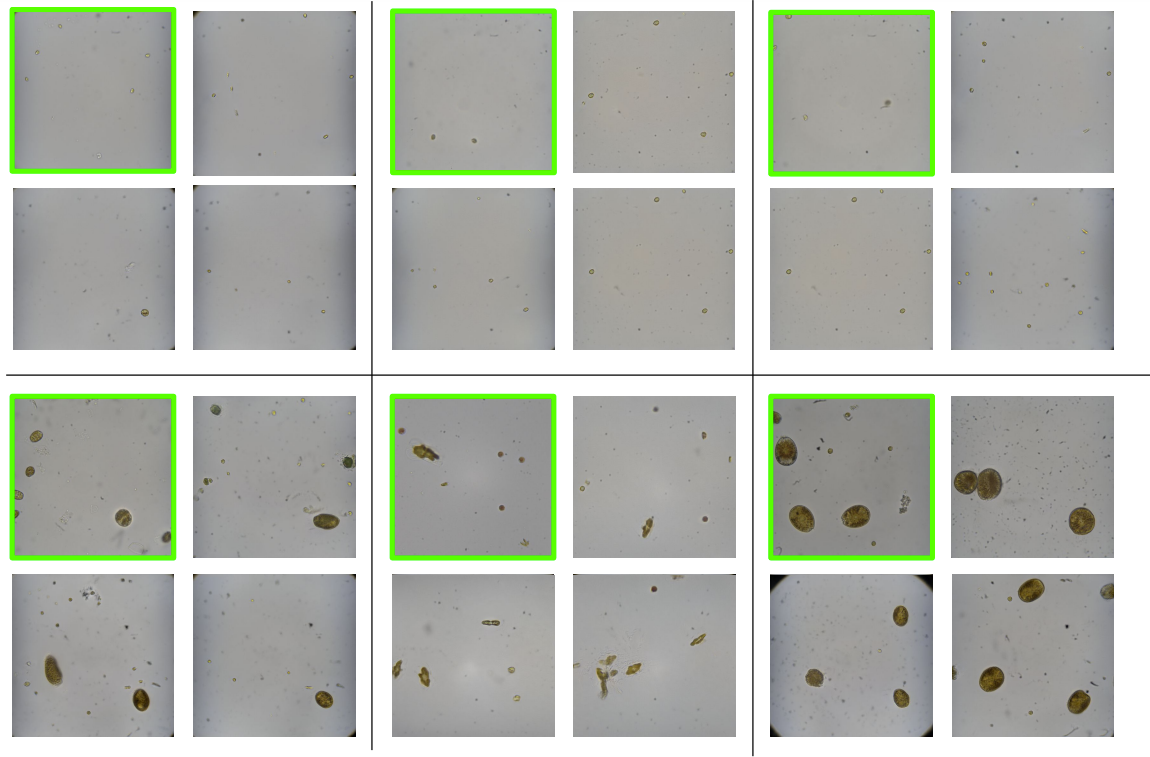}
        \caption{More examples of Nearest Neighbors : the top left image (in green) in each box is the generated image, the rest three images are its nearest neighbours in the real dataset in pixel space (top row), and in feature space (bottom row).}\
\end{figure}

\end{document}